# Risk-Informed Diffusion Transformer for Long-Tail Trajectory Prediction in the Crash Scenario

Junlan Chen[1, 2*], Pei Liu[3*], Zihao Zhang[1], Hongyi Zhao[1], Yufei Ji[1], Ziyuan Pu[1]

[1]School of Transportation, Southeast University, No.2 Southeast University Road, Nanjing, China, 211189. E-mail: chenjunlan@seu.edu.cn; zihaozhang@hnu.edu.cn; hongyi.zhao@seu.edu.cn; jiyufei@seu.edu.cn; ziyuan.pu@monash.edu

[2]School of Civil Engineering, Monash University, Melbourne, Australia. E-mail: junlan.chen@monash.edu;

[3]Intelligent Transportation Thrust, Systems Hub, The Hong Kong University of Science and Technology (Guangzhou), Guangzhou, China. E-mail: pliu061@connect.hkust-gz.edu.cn

*These authors contributed equally to this work

**ABSTRACT**

Trajectory prediction methods have been widely applied in autonomous driving technologies. Although the overall performance accuracy of trajectory prediction is relatively high, the lack of trajectory data in critical scenarios in the training data leads to the long-tail phenomenon. Normally, the trajectories of the tail data are more critical and more difficult to predict and may include rare scenarios such as crashes. To solve this problem, we extracted the trajectory data from real-world crash scenarios, which contain more long-tail data. Meanwhile, based on the trajectory data in this scenario, we integrated graph-based risk information and diffusion with transformer and proposed the Risk-Informed Diffusion Transformer (RI-DiT) trajectory prediction method. Extensive experiments were conducted on trajectory data in the real-world crash scenario, and the results show that the algorithm we proposed has good performance. When predicting the data of the tail 10\% (Top 10\%), the minADE and minFDE indicators are 0.016/2.667 m. At the same time, we showed the trajectory conditions of different long-tail distributions. The distribution of trajectory data is closer to the tail, the less smooth the trajectory is. Through the trajectory data in real-world crash scenarios, Our work expands the methods to overcome the long-tail challenges in trajectory prediction. Our method, RI-DiT, integrates inverse time to collision (ITTC) and the feature of traffic flow, which can predict long-tail trajectories more accurately and improve the safety of autonomous driving systems.

Risk-Informed Diffusion Transformer for Long-Tail Trajectory Prediction in the Crash Scenario

**INTRODUCTION**

The emergence of autonomous driving is a revolutionary technical progress set to transform transport fundamentally. Within this context, trajectory prediction plays a pivotal role, focusing on forecasting the future paths of road agents based on their historical movements. Accurate predictions of surrounding agents' trajectories enable autonomous vehicles to strategically plan their routes and avert potential collisions. Contemporary research has introduced a multitude of trajectory prediction methodologies, encompassing both unimodal and multimodal approaches. These models usually face two challenges to achieve high prediction accuracy.

First is the long-tailed phenomenon in the datasets used to train the models. Long-tailed phenomenon refers to the lack of certain types of data while rich in other types of data (*1*). In real traffic scenarios, most of the drivers make smooth maneuvers, which can be easily predicted. This makes the average error of the model seem small but actually ignores many key scenarios, such as crashes. Crash scenarios contain abundant long-tail data, in which agents make more lane changes, sharp turns, and dodges. However, crashes are sporadic and low-probability events, and the existing datasets lack trajectory data of agents in crash scenarios. Therefore, models trained based on these datasets usually have poor performance in critical scenarios, which can cause severe traffic crashes (*2, 3*).

Another challenge is the lack of authenticity in generating trajectories. Recently, there was a trend for generating trajectories by generative adversarial networks (GANs) (*4*) and Variational Auto-Encoders (VAEs) (*5*). However, although the behaviors generated by these models seem reasonable, they only mirror the distribution of the training data and do not learn any mechanism of traffic flow. In the meantime, these methods ignored vehicle motion uncertainty caused by different driving styles. Moreover, the interrelationship between the vehicle and the environment was rarely considered (*6*). The limitations of these methods lead to less practical trajectory prediction in the real world (*7*).

In order to address the problem of long-tail phenomenon, typical methods are class rebalancing, information augmentation, and module improvement (*3, 8, 9*). Most model improvement methods are based on class-rebalancing. However, class-rebalancing methods have the drawback of improving tail-class performance at the cost of lower head-class performance (*8*). This will reduce the overall accuracy of the model. Then, existing information augmentation methods cannot utilize information of multiple dimensions well. However, many types of research have demonstrated the important role of risk-information and traffic flow characteristics (flow, flow rate, and flow velocity) in trajectory prediction, especially in complicated crash scenarios (*3, 9*). So, we introduced graph-based risk information and traffic flow information into our prediction model to improve its performance.

More recently, diffusion probabilistic models, also known as diffusion models, have shown more authentic results in trajectory generation and are considered to be a model with great potential (*10*). Diffusion probabilistic models, which introduce noise in the forward diffusion process and then denoise to reconstruct the original trajectory distribution, have become a highly adaptable model for trajectory generation (*11, 12*). To further enhance the performance of diffusion probabilistic models, recent studies have improved this approach by replacing the traditional convolutional U-Net with Transformers, which have shown a more effective perception of multi-dimensional information in trajectory modeling (*13*). Therefore, we adopt the diffusion with transformers (DiT) as a trajectory generator to produce more realistic trajectories.

In this paper, we utilized the trajectory data of a crash scenario to enhance the model's prediction ability in the long-tail datasets. To enhance the accuracy of trajectory prediction in crash scenarios, we introduced risk information into the prediction, including inversed time to collision (ITTC), velocity, and traffic flow features. A diffusion model with Transformer is utilized to generate the feature of predicted trajectories. Our contribution can be summarized as follows:

- **RI-DiT:** We proposed a new method to enhance the accuracy of trajectory prediction in crash scenarios, which is called Risk-informed Diffusion with Transformers (RI-DiT). By utilizing the DiT module to generate multi-modal trajectories, the diversity of the predicted trajectories is enriched, making the predictions more realistic.

- **Graph-based risk information:** We extracted the trajectories in the crash scenarios to enrich the long-tail data and improved the model's prediction performance for long-tail data by integrating risk-info and traffic flow in DiT. Through diversified input data, the prediction performance of the model in crash scenarios is improved.

- **Outperformance:** We conducted extensive experiments on our algorithm using the realworld vehicle trajectory in a crash scenario. The results demonstrate that our proposed algorithm performs exceptionally well, achieving notably values for the minimum average displacement error (minADE) and minimum final displacement error (minFDE) metrics, with results of 0.009/1.519 m, respectively.

## RELATED WORKS

### Trajectory Prediction

Many methods have been proposed to predict future trajectories, including early physics-based models, such as Kalman filter (*14*) and Monte Carlo Methods (*15*). With the development of neural networks, deep learning methods are proposed to predict future trajectories, such as Recurrent Neural Network (RNN) (*16*), which stores information of the previous timesteps and determines outputs with the information of the contemporary inputs and history information. However, it has a drawback of gradient explode or decay when the timesteps are large. Intergrades of RNN, LSTM, and gated recurrent unit (GRU) were proposed to solve the problem of gradient explosion or decay. In practice, they are used as sequence models to predict vehicle maneuvers. In (*17*), fully connected layers were applied to extract latent features, and LSTM layers were then used to predict future trajectory. In (*18*), an encoder-decoder LSTM structure architecture was used where an LSTM encodes the history trajectory features, and an LSTM decoder solves the most likely future trajectories. However, given a partial history, there is no single correct prediction of the future, and these approaches cannot cope with the multimodal nature of the future.

Multiple trajectories are plausible and socially acceptable. Hence, multimodal trajectory prediction methods are essential as they can generate a diverse set of possible future trajectories, capturing the full range of plausible outcomes. To generate multimodal trajectories, Goal-GAN (*19*) first estimates possible goal positions and then uses RNN to generate trajectories towards these goals. Another model (*20*) uses a Conditional Variational Autoencoder (CVAE) to generate a multimodal trajectory distribution by utilizing the Gaussian Mixture Model, effectively capturing the diversity and uncertainty in human behavior. In conclusion, the model must encode a variety of action possibilities into a latent space and utilize decoders to generate diverse and realistic future paths in order to facilitate the generation of multimodal trajectories.

To generate more realistic and diverse trajectories, many researchers enhance their models with physical laws or other knowledge. For instance, Pi-GAN (*21*) incorporated a GAN model with a physics-informed discriminator. In PIT-IDM (*22*), a diffusion model is enhanced by physics messages. However, in crash scenes, vehicles make sharp turns, leading to higher accelerations that do not follow physical constraints and making them unsuitable for using physical laws to inform the model. Conversely, risk information is widely used for crash-scene prediction. For example, one study (*3*) utilized traffic parameters such as flow, density, and speed, and various surrogate safety measures (SSM), such as TTC, to predict the possibility of collision. SSMs have proven to be highly important in pre-crash scenes (*23*). Moreover, traditional diffusion probabilistic models inherited convolutional U-Nets from CNN++ with a few changes (*13*). Transformer, with its attention mechanism, has demonstrated superior performance compared to U-Nets in processing information more efficiently and accurately. For example, TrajFormer (*24*) leverages Continuous Point Embedding and Squeezed Transformer Blocks to effectively capture spatio-temporal dependencies in irregular trajectory data while significantly improving computational efficiency. TUTR (*25*) unifies trajectory prediction components into a transformer architecture, eliminating the need for post-processing while significantly improving prediction accuracy and inference speed.

**Long Tail Learning**

To address the problem caused by the long-tail phenomenon, resampling has been an intuitive choice (*26–28*). The random under-sampling (RUS) method, which involves the random exclusion of data from the head classes, has the inherent disadvantage of impairing the performance of the model on head classes (*29*). Progressively-balanced sampling (*30*) combined random sampling with class-balanced sampling. However, the dataset is not necessarily classified, making it hard to know the frequency of each class of data. Moreover, these methods improve its performance on long-tail data at the cost of its performance on head data (*8*).

Recently, the generative adversarial model (GAN) (*31, 32*) has emerged as a promising way to generate more data for long-tail scenes. For example, the LSTM-GAN (*33*) model utilizes long-short term memory (LSTM) (*17*) layers to capture the temporal dynamics involved in traffic flows. However, mode collapse is one of the most common problems with GANs. When mode collapse occurs, the generator tends to produce only a few types of samples, ignoring other modes of the data distribution (*34*). Additionally, the training process for GANs can be unstable due to adversarial learning.

Unlike GAN, VAE is a type of explicit generative model that models probability density $p(x)$ explicitly and is widely used in prediction works. For instance, TrajVAE (*35*) utilized VAE to generate trajectories, which is proved to outperform TrajGAN in terms of accuracy and stability. It achieves better similarity scores (Hausdorff and Dynamic Time Warping distances) and is more robust in generating realistic trajectories. However, VAEs assume that the latent variables follow a standard Gaussian distribution, which may not always align with the actual distribution of the data. Moreover, these methods focus on generating only one path for vehicles, neglecting that there may be multiple paths for the vehicle.

To address the above issues, we used the most contrastive long-tail data of a collision scenario to improve the predictive ability of the model. In order to increase the authenticity and variance of the generated trajectory, we introduced risk information into the prediction, including time to collision (TTC), velocity, and the feature of traffic flow (*36*). A diffusion model with transformer is used to generate the predicted trajectories. Additionally, we used multi-layer perceptron (MLP) layers to decode agent latent into multiple possible future trajectories to generate multimodal trajectories.

## METHOD

Figure 1 shows our method framework. By inputting traffic flow characteristics, trajectory characteristics, and graph-based risk information into the risk conditional diffusion module and fusing the risk information by adding random noise to generate trajectory features. The trajectory features are passed into the scene encoder for encoding, and then the multimodal trajectories are generated through the trajectory decoder. The part of the method details the methods in the framework.

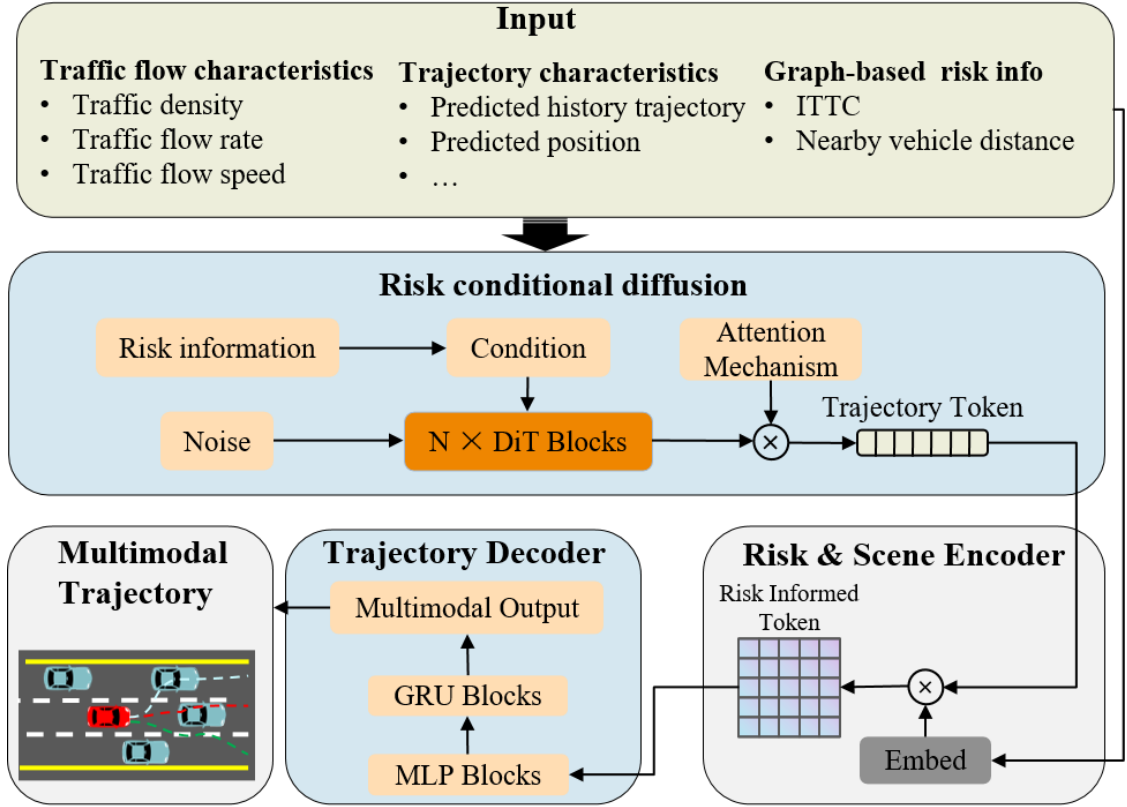

FIGURE 1 Overall Architecture of the Method. We extracted vehicle trajectory features from crash scenarios and generated multimodal Trajectory through three modules: Risk Conditional Diffusion, Risk & Scene Encoder, and Trajectory Decoder.

## A Graph-based Risk Calculation

We established a graph theory-based method (*37, 38*) to explicitly calculate the risk information of a certain vehicle. The risk network is defined as an undirected graph $G = (V, E, W)$, in which $V = v_1, v_2, ..., v_n$ is a set of vertices, each representing a vehicle at the current scene. $E \in \{<v_a, v_b>: v_a, v_b \in V, a \models b\}$ is a set of edges that each connects two vertices. At the same time, we used ITTC (*39*) as the weight between two vertices. ITTC avoids the issue of TTC being infinite when two vehicles are moving away from each other. Figure 2 shows the positional relationship between the two cars, each vehicle has a specific position, $O_j : (x_j, y_j)$ and velocity, $V_j : (v_x{}^j, v_y{}^j)$.

ITTC assumes the two vehicles keep their speed and direction before a potential crash, and

$$ITTC_{ij} = \begin{cases} \dfrac{v_x^i - v_x^j}{x_i - x_j - (L_i + L_j)/2} & (v_x^i - v_x^j)[x_i - x_j - (L_i + L_j)/2] > 0 \\ 0 & (v_x^i - v_x^j)[x_i - x_j - (L_i + L_j)/2] < 0 \end{cases}$$

(1)

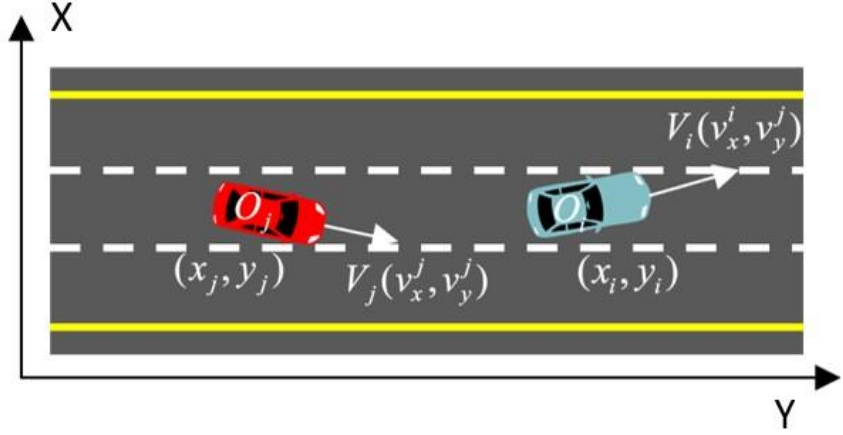

FIGURE 2 Schematic Diagram of the Positional Relationship between Two Vehicles. can be calculated by the Equation (1).

Where $L_i$ and $L_j$ respectively represent the length of the vehicle. $y_i$ and $y_j$ respectively represent the positions of vehicle i and vehicle j in the Y-axis direction, and $v_y^i$ and $v_y^j$ respectively represent the velocities of vehicle i and vehicle j in the Y direction.

If two vehicles are approaching in the lateral direction, replace longitudinal speed, longitudinal distance, and length with lateral speed, lateral distance, and width.

$$ITTC_{ij} = \begin{cases} \dfrac{v_y^i - v_y^j}{y_i - y_j - (L_i + L_j)/2} & (v_y^i - v_y^j)[y_i - y_j - (L_i + L_j)/2] > 0 \\ 0 & (v_y^i - v_y^j)[y_i - y_j - (L_i + L_j)/2] < 0 \end{cases}$$

(2)

In this paper, 6 vehicles are taken into consideration, such as Figure 3, which are the nearest vehicles in the front and rear on the nearest three lanes. We used weighted degree centrality to estimate the risk of certain vertices, as shown in the following equation:

$$C_d(v_i) = \sum_{j=1, j \neq i}^{n} w_{e_{ij}}$$

(3)

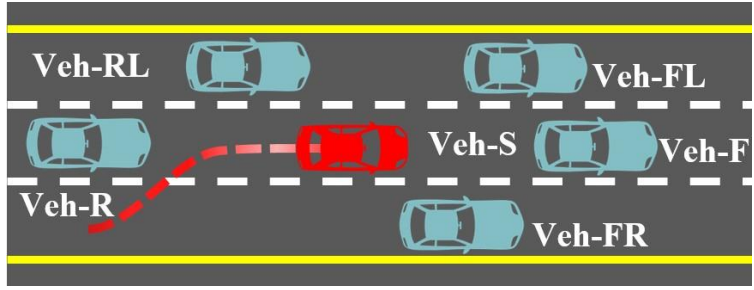

FIGURE 3 Schematic Diagram of the Vehicle Scenario Selected for the Experiment.

**Trajectory Diffusion**

In crash scenarios, agents make more lane changes, sharp turns, and dodges, making the trajectories less smooth than ordinary ones. Moreover, more indicators such as risk, flow, and mean velocity are needed to be taken into consideration. By introducing more uncertainty into the trajectory generation process, we used the DiT module to encode trajectories into latent space to generate more diverse multimodal trajectories in crash scenarios (*12*). Traditionally, a denoising diffusion probabilistic model (DDPM) inherits convolutional U-Net from CNN++, the DiT module means that U-Net replaced by Transformer (*13*), which outperforms conventional U-Net. Transformer (*40*) is a deep learning model initially designed for natural language processing that uses selfattention mechanisms and parallel processing to efficiently handle long sequences of data, widely used in tasks like translation and text generation. DDPM consists of two phases: forward process and reverse process (denoising process). Given a set of trajectory data $x_0$ $q(x_0), x_1, x_2, .... x_T$ are latent of the same dimension, $x_0^S$ is the synthetic trajectory in the forward process, according to a variance schedule $\beta_1, \beta_2, ..., \beta_T$ in each timestep, Gaussian noise $N(\cdot)$ is added into the data.

$$q(x_{1:T} \mid x_0) := \prod_{t=1}^{T} q(x_t \mid x_{t-1}); \quad q(x_t \mid x_{t-1}) := \mathcal{N}(x_t; \sqrt{1-\beta_t} x_{t-1}; \ \beta_t \mathbf{I})$$

(4)

Where $\beta_t$ is a learnable parameter, $x_t$ can be expressed as $x_t = \bar{\alpha}_t x_0 + \sqrt{1-\bar{\alpha}_t}\varepsilon$, and $\varepsilon \sim N(0, I)$, $\bar{\alpha}_t = \prod_{i=1}^{t}(1-\beta_i)$. The reverse process can be defined as a Markov chain starting at $p(x_T) = N(x_T; 0, I)$, which aims to learn the joint distribution $p_\theta(x_{0:T})$ and recover the trajectory from noises.

$$p^\theta(x_{0:T}) := p(x_T) \prod_{t=1}^{T} p_\theta(x_{t-1} \mid x_t), p_\theta(x_{t-1} \mid x_t) := \mathcal{N}(x_{t-1}; \mu_\theta(x_t, t), \Sigma_\theta(x_t, t))$$

(5)

where $\mu_\theta(x_t, t) = \frac{1}{\sqrt{\alpha_t}}(x_t - \frac{\beta_t}{\sqrt{1-\bar{\alpha}_t}}\epsilon_\theta(x_t, t))$, $\tilde{\beta}_t = \frac{1-\bar{\alpha}_{t-1}}{1-\bar{\alpha}_t}\beta_t$, and $\Sigma_\theta(x_t, t) = \tilde{\beta}_t^{\frac{1}{2}}$

The core of training the diffusion model is to minimize the mean squared error between the Gaussian noise $\varepsilon$ and predicted noise level, as shown in:

$$L(\theta) := \mathbb{E}_{t, x_0, \epsilon} \left[ \| \epsilon - \epsilon_\theta (\sqrt{\overline{\alpha}_t} \, x_0 + \sqrt{1 - \overline{\alpha}_t} \, \epsilon, t) \|^2 \right]$$

(6)

Where $\overline{\alpha}_t$ is a hyperparameter, $\epsilon_\theta$ is a function that takes $x_t$ to predict $\epsilon$, and $\theta$ is a learnable parameter.

Figure 4 shows the architecture of the conditional denoising diffusion probabilistic model (DDPM), which takes random noise, history trajectory, and condition as input. The output is predicted noise, and the action is latent. In our model, risk information is included as the condition. This enables the model to acquire more information about crash scenarios and enhance the model's predictive ability in critical scenarios.

**Scene Encoder**

In crash scenes, there is more information needed to be taken into consideration. We encode trajectory, risk information, and traffic flow information into tokens by using different embedding blocks. We encode trajectory, risk information, and traffic flow information into tokens by using different embedding blocks. In the process of encoding history trajectory *traj$_i$* into a unified onedimensional token. So that the trajectory doesn't have to be converted to a Frenet coordinate

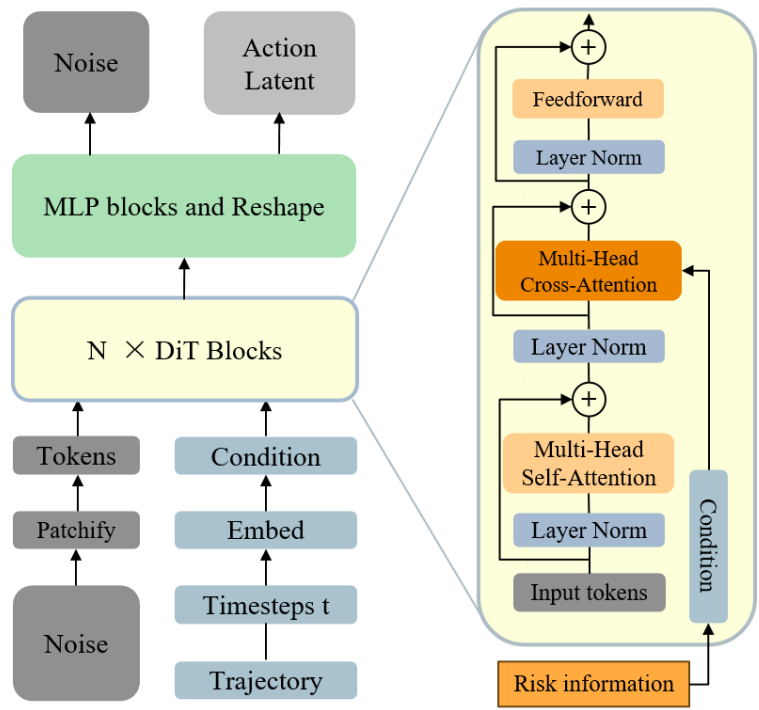

FIGURE 4 The Architecture of Conditional DDPM. The left is the overview of DDPM, which takes trajectories and noises as inputs and outputs Noise and action latent. The right is the DiT Blocks framework, which integrates risk informed through multi-head cross-attention.

system. For a given vehicle *i* to be predicted:

$$E_{traj} = \phi_{traj}[x_i, y_i] \tag{7}$$

Where $\varphi_{traj}$ represents a linear layer that transforms trajectory information into a token, and $E_{traj}$ represents the result of trajectory embedding.

The risk embedding outcome $E_{risk}$ is defined as:

$$E_{risk} = \phi_{risk}[w_1, w_2, \ldots, w_6] \tag{8}$$

Where $w_i$ are the weights of the 6 vehicles taken into consideration. $\varphi_{risk}$ is a linear layer that transforms risk information into a token.

The traffic flow characteristics are also turned into tokens:

$$E_{env} = \phi_{env}[K_{env}, Q_{env}, V_{env}] \tag{9}$$

Where $K_{env}$, $Q_{env}$, $V_{env}$ are the flow, flow rate, and mean velocity of the traffic flow.

The driver's driving characteristics are also processed:

$$E_v = \phi_v[a_v, m_v, std_v, k_v, skew_v, cv_v] \tag{10}$$

Where $a_v$, $m_v$, $std_v$, $skew_v$ are the acceleration, average value, standard deviation, kurtosis, skewness, and coefficient of variation of the velocity.

To synthesize these tokens into one token $E_A$, we concatenate $E_{traj}$ and $E_{risk}$ into one final token:

$$E_A = \mathrm{Re}\, LU(LayerNorm(Concat(E_{traj}, E_{risk}, E_{env}, E_v))) \tag{11}$$

Where $ReLU(\cdot)$ is the activation function (*41*).

In this process, we have a comprehensive token that contains information on both history, trajectory, and risk. The token is then fed to DiT blocks, which utilize a transformer to provide precise information. The transformer is built with layers including multi-head self-attention mechanisms and feed-forward neural networks, enhanced by positional encoding and residual connections. This architecture enables efficient parallel processing of sequences, making it powerful for tasks like translation and text generation.

**Trajectory Decoder**

The trajectory decoder decodes the risk-informed token and trajectory token into predicted future trajectory. In this model, GRU layers (*42*) and MLP layers are applied to predict future trajectories. The GRU blocks uses a reset gate and an up-date gate, which help control the flow of information and mitigate the vanishing gradient problem to predict the future risk-informed token. The predicted information is then fed to MLPs to decode into displacements of multimodal future timesteps. The trajectory in predicted timestep m is then formed as:

$$Traj_m = Pos_t + \sum_{i=t}^{T_f}[\Delta x_m, \Delta y_m, \Delta \theta_m] \tag{12}$$

Where $Pos_t$ is the last history position of the vehicle, and $[\Delta x_m, \Delta y_m, \Delta \theta_m]$ are the displacements of $x, y$ position and angle of trajectory m.

The velocity can be calculated by following equation:

$$v_m = \frac{[\Delta x_m, \Delta y_m]}{\Delta t} \tag{13}$$

Where $\delta t$ is the time difference between the two consecutive points.

## EXPERIMENTS

### Dataset

The dataset we adopted is the public dataset released by the I-24MOTION project located near Nashville, Tennessee, USA (*43*). This project consists of 276 high-resolution traffic cameras, and these cameras are distributed along approximately 4.2 miles of Interstate-24 in the USA. To collect sufficient data under key scenarios, we selected the vehicle trajectories near section MM59.7 on November 21, 2022. On that day, a rear-end collision occurred among nearby vehicles, causing traffic congestion from 6:14 to 7:43 AM. The trajectory data in this crash scenario may generate more abnormal trajectories due to the traffic congestion caused by the crash. To better explore the relationship between vehicle trajectories and surrounding vehicles, we only selected the scenarios of the six closest vehicles.

The main data information we used is shown in Table 1. To measure the risk information and the traffic flow data in crash scenarios, we calculated the ITTC value in each frame scenario and the traffic flow density, speed, and flow rate in the road section at that time. Then, the features such as the historical trajectories, speeds, distances to surrounding vehicles, and traffic flow of the vehicles were taken as input features to predict the vehicle trajectories in the next 50 frames. Meanwhile, We calculated the FDE of each trajectory using a simple model in order to distinguish the long-tail data. To ensure there are sufficient long-tail data in the test set, we first divided the non-long-tail data in a ratio of 2:8 and then added the long-tail data to the training set and the test set in a ratio of 5:5 respectively. This made the amount of long-tail data in the training set and the test set each account for 10% of the total data. Therefore, the actual ratio of the training set to the test set is 74:26.

TABLE 1 Summary of Variables Description (Count = 16, 960).

| Variables | Description |
| --- | --- |
| Trajectory characteristics | |
| Predicted history trajectory | A trajectory sequence consisting of 75 trajectory points (x, y). |
| Predicted position | The starting position of the predicted trajectory of the vehicle. |
| Δ Predicted history trajectory | Increment of historical trajectory coordinates. |
| Velocity in the x-direction | The velocity of the vehicle in the x-axis direction (mean = 4.06 km/h). |
| Velocity in the y-direction | The velocity of the vehicle in the y-axis direction (mean = 0.26 km/h). |
| FDE | The FDE predicted by the simple model is used to divide the long-tail data (mean = 4.16m). |
| Risk information | |
| ITTC | The reciprocal sequence of the collision time between the vehicle and the surrounding vehicles at each point. |
| Nearby vehicle distance | The distance of the surrounding vehicles. To prevent the distance from being infinite, it is expressed by the reciprocal. |
| Traffic flow characteristics | |
| Traffic flow density | The traffic flow density of the same road section at the same moment (mean = 51.83 Veh/km). |
| Traffic flow rate | The traffic flow rate of the same road section at the same moment (mean = 1042 Veh/h). |
| Traffic flow speed | The traffic flow speed of the same road section at the same moment (mean = 4.38 km/h). |
| Target variable | |
| Predicted future trajectory | The trajectory coordinates to be predicted consist of 50 trajectory points. |

## Metrics and Loss Function

Metrics. The performance metrics we adopted are minADE and minFDE, and these two trajectory prediction evaluation metrics have been widely used in multi-trajectory modality research (*1, 44, 45*). Among them, average displacement error (ADE) calculates the average $L_2$ distance between all predicted trajectories and the actual trajectories and reports the error value that is closest to the actual trajectory. The final displacement error calculates the $L_2$ distance between the final position of the predicted trajectory and the final position of the actual trajectory and reports the minimum value. To measure the algorithm's performance, we take the average of the minADE and minFDE, which are calculated for all trajectories.

Long-Tail Sample Selecting. To better verify the prediction performance of the algorithm for the long-tail trajectory data in crash scenarios, we first predicted the trajectories through the relatively simple LSTM model (*46*), calculated the FDE value of each trajectory, and used this to evaluate the difficulty of trajectory prediction. Next, all FDEs were divided into 100 parts with the interval of [0.0002, 41.3775] through the frequency histogram shown in Figure 5. It shows that when the trajectories belong to the regular trajectories (head data), the prediction accuracy is relatively high. However, when the data are distributed in the tail (long-tail data), the prediction accuracy is low. To compare the performance ability of the model in the long-tail data, we divided these 100 data into six grades: Top 10%, Top 20%, Top 30%, Top 40%, Top 50%, and Rest.

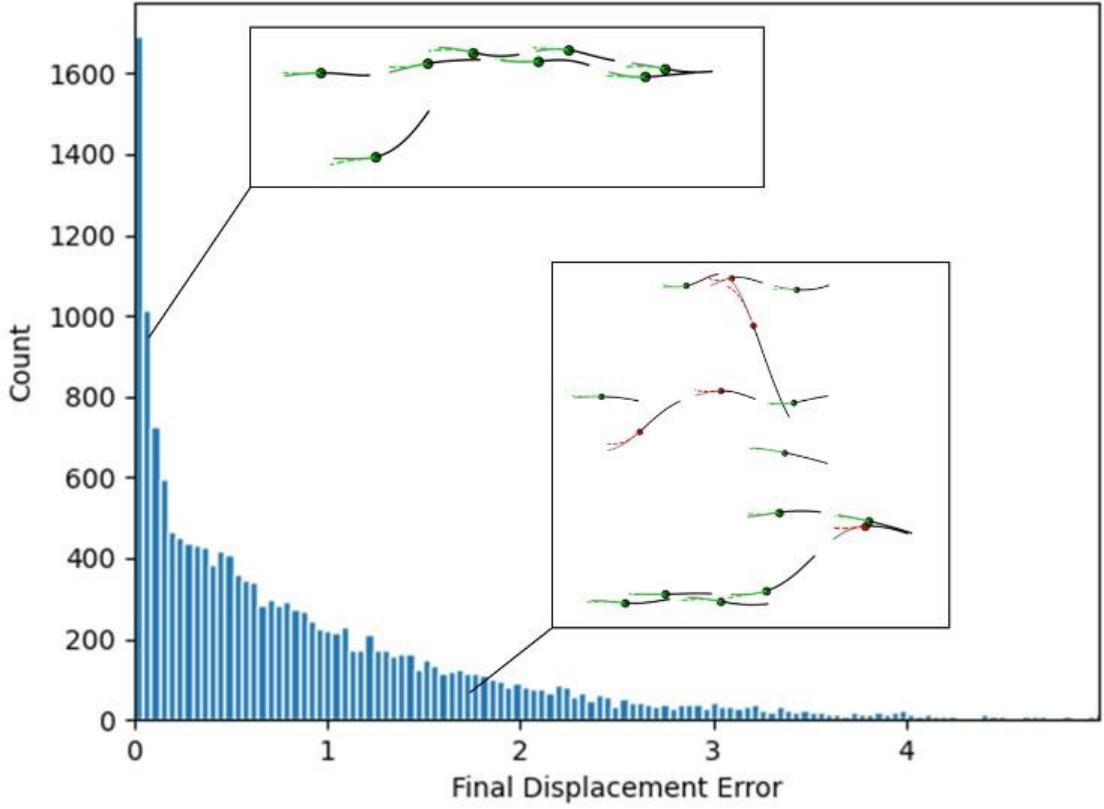

FIGURE 5 Histogram of FDE Distribution for Trajectory Dataset. The rectangular boxes in the chart represent the trajectory prediction under this distribution, the black dots indicate the current position of the vehicle, the green dashed lines represent accurately predicted trajectories and the red dashed lines represent inaccurately predicted trajectories.

Loss function. The objective is to forecast the future, diverse, and realistic trajectory of a specific vehicle. To this end, we propose the incorporation of a Huber loss function (*47*) between the predicted trajectory and the ground truth. The trajectory prediction loss is defined as:

$$L_{Traj} = \begin{cases} \frac{1}{2}(Traj_{pred} - Traj_{gt})^2 & |Traj_{pred} - Traj_{gt}| < \delta \\ \delta|Traj_{pred} - Traj_{gt}| - \frac{1}{2}\delta^2 & otherwise \end{cases} \tag{14}$$

Where $\delta$ is a hyperparameter that determines the threshold of using MSE or MAE as the loss function. $Traj_{pred}$, and $Traj_{gt}$ are the predicted trajectory and future ground truth.

In conclusion, the total loss function is the combination of $L_{Traj}$ and $L_{diff}$, and is defined as:

$$L_{total} = L_{diff} + L_{Traj} \tag{15}$$

Where $L_{diff}$ is the loss of DiT module, which is mentioned in the part of trajectory diffusion.

Experiment Setting

The experimental equipment we adopted is one NVIDIA A800 GPU, and Adam is used as the model optimizer (*48*). Based on the results of existing studies and the characteristics of our data, we set the batch size of the training network to 128 and the number of training rounds (epochs) to 100. Referring to the model parameter settings of WcDT (*10*), we set the learning rate to $1 \times 10^{-4}$, the diffusion time steps to 50 frames, and the dropout rate to 0.1. The framework we provided mainly consists of four parts: DiT, scene encoder, trajectory decoder, and multimodal trajectory generation. The specific settings of these modules are as follows: DiT adopts the Gaussian diffusion process, the scene encoder uses a 3-layer MLP to encode the vehicle position, vehicle speed features, ITTC features, and the distances to surrounding vehicles, the number of MLP layers of the trajectory decoder is set to 2, and the number of multi-modal trajectory outputs is set to 10.

**Comparative Results**

To better quantify the performance of the model, we conducted a series of comparative experiment analyses. Table 2 shows the effect of the model when different modules are changed. This result indicates that the DiT module and the multimodal module can positively reduce the model's positive error in trajectory prediction. Meanwhile, from the data results of different distributions, the prediction of the top 10% long-tail trajectories is the most difficult, with minADE and minFDE being 0.016 m and 2.676 m, respectively. As the data distribution changes from the tail to the head, the prediction difficulty gradually decreases. However, it can be seen that we also maintained a good prediction effect on the long-tail data. The differences between the head prediction result and the tail prediction result in minADE and minFDE are 0.010 m and 1.905 m, respectively.

TABLE 2 The Impact of Different Modules on the Validation Dataset.

| Multi-modal | DiT | Metrics | Top 10% | Top 20% | Top 30% | Top 40% | Top 50% | Rest | All |
|---|---|---|---|---|---|---|---|---|---|
| | | minADE | 0.022 | 0.019 | 0.017 | 0.017 | 0.016 | 0.008 | 0.013 |
| | | minFDE | 2.747 | 2.380 | 2.285 | 2.195 | 2.097 | 0.788 | 1.586 |
| | ✓ | minADE | 0.021 | 0.018 | 0.017 | 0.017 | 0.016 | 0.008 | 0.013 |
| | ✓ | minFDE | 2.861 | 2.455 | 2.331 | 2.237 | 2.146 | 0.808 | 1.493 |
| ✓ | | minADE | 0.020 | 0.017 | 0.017 | 0.015 | 0.015 | 0.007 | 0.012 |
| ✓ | | minFDE | 2.717 | 2.419 | 2.361 | 2.208 | 2.083 | 0.814 | 1.579 |
| ✓ | ✓ | minADE | 0.016 | 0.014 | 0.013 | 0.013 | 0.012 | 0.006 | 0.009 |
| ✓ | ✓ | minFDE | 2.676 | 2.269 | 2.211 | 2.093 | 1.981 | 0.771 | 1.519 |

The Trajectory Decoder is responsible for decoding the features and generating multimodal trajectories. The performance of this module determines the result of trajectory prediction. To prevent overfitting, we compared the number of MLP linear layers. In Table 3, we can see that for our data, setting the MLP linear layer to 2 is the most appropriate. Setting too many MLP linear layers will not only lead to model overfitting but also increase the number of parameters, which is not conducive to the real-time reasoning calculation of the model. At the same time, it can be seen that the prediction results of different model complexities in different data distributions are also different. When MLP is set to 4, the minFDE is reduced by 0.261 m in Rest data compared to when MLP is set to 2 but is reduced by 0.426 m in Top10% data.

TABLE 3 The Impact of Different MLP Layers in the Trajectory Decoder on the Validation Dataset.

| Decoder layer | Metrics | Top 10% | Top 20% | Top 30% | Top 40% | Top 50% | Rest | All |
|---|---|---|---|---|---|---|---|---|
| 8 | minADE | 0.153 | 0.131 | 0.125 | 0.119 | 0.114 | 0.047 | 0.088 |
| | minFDE | 7.827 | 6.631 | 6.338 | 6.061 | 5.795 | 2.357 | 4.459 |
| 4 | minADE | 0.030 | 0.026 | 0.025 | 0.023 | 0.022 | 0.006 | 0.016 |
| | minFDE | 2.250 | 1.923 | 1.831 | 1.737 | 1.607 | 0.510 | 1.194 |
| 2 | minADE | 0.016 | 0.014 | 0.013 | 0.013 | 0.012 | 0.006 | 0.009 |
| | minFDE | 2.676 | 2.269 | 2.211 | 2.093 | 1.981 | 0.771 | 1.519 |

One of the main features of this paper is to predict the long-tail trajectory data in crash scenarios by generating multimodal trajectories. Therefore, we conducted comparative experiments for different numbers of modalities. Table 4 shows the comparison effects of different modalities. The results show that as the number of modalities increases, minADE and minFDE tend to be the minimum values. But from the results, it can be seen that when multimodal is set to 10, it is only a smaller value than when it is set to 30. Taking the Rest of the data with the largest difference as an example, the prediction error only increases by 0.001/0.009 m. Although the improvement in effect is not obvious, due to the generation of more trajectories, the amount of data has increased by 13.8

TABLE 4 Comparison of Model Effects Corresponding to Different Modalities.

| Multimodal | Metrics | Top 10% | Top 20% | Top 30% | Top 40% | Top 50% | Rest | All |
|---|---|---|---|---|---|---|---|---|
| 30 | minADE | 0.016 | 0.014 | 0.013 | 0.012 | 0.011 | 0.005 | 0.009 |
| | minFDE | 2.562 | 2.261 | 2.147 | 2.061 | 1.969 | 0.762 | 1.497 |
| 10 | minADE | 0.016 | 0.014 | 0.013 | 0.013 | 0.012 | 0.006 | 0.009 |
| | minFDE | 2.676 | 2.269 | 2.211 | 2.093 | 1.981 | 0.771 | 1.519 |
| 1 | minADE | 0.021 | 0.018 | 0.017 | 0.017 | 0.016 | 0.008 | 0.013 |
| | minFDE | 2.861 | 2.455 | 2.331 | 2.237 | 2.146 | 0.808 | 1.493 |

To compare the effects of trajectory prediction in different data distributions, we selected representative trajectories in each data distribution. As shown in Figure 6, when the data distribution is located at the tail of the Top 10%, the vehicle trajectory changes greatly, making it difficult to predict the trajectory. Multiple trajectories can be generated through multimodal trajectories, and the trajectory with the highest probability can be obtained by calculating the confidence level. When the data distribution is located at the Rest, the vehicle trajectory is relatively smooth and easy to predict. However, multimodal trajectories still consider the possibility of vehicle turning and generate trajectory predictions in other directions. This figure, on the one hand, shows the vehicle movement trajectories in various data distribution scenarios, and on the other hand, also shows the prediction effect of our method and the advantages of multimodal trajectory prediction for long-tail scenarios.

## DISCUSSION

The long-tail trajectories in crash scenarios contain many driving behaviors taken for emergency avoidance, but current studies pay less attention to the prediction of this type of trajectory, which leads to autonomous vehicles predicting incorrect trajectories in some critical scenarios, thereby causing serious traffic crashes (*49*). To address this problem, we selected the vehicle trajectory data in the rear-end crash scenario from the I-24MOTION public dataset (*43*). This scenario contains some trajectory data of the approximately one-and-a-half hours of traffic congestion caused by the crash. Based on the trajectory data, the risk information in the scene is obtained through graph-based risk calculation. Incorporating such data containing risk information can improve the trajectory prediction ability of intelligent driving vehicles in emergency avoidance scenarios, thereby reducing the occurrence of crashes.

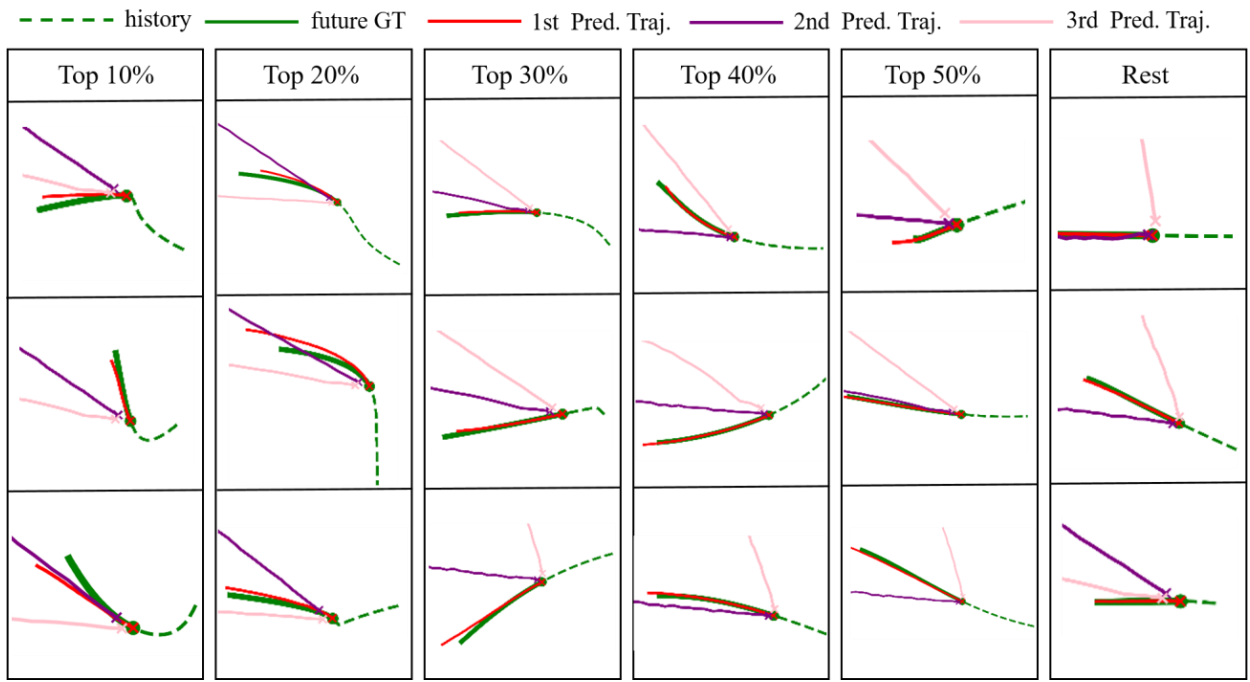

FIGURE 6 Illustration of Predictive Results for Different Distributions of Modal Data. GT denotes the ground truth trajectory, 1st Pred. Traj. represents the prediction with the highest confidence, 2nd Pred. Traj. indicates the prediction with the second highest confidence, and 3rd Pred. Traj. signifies the prediction with the third highest confidence.

For the trajectory data in the crash scenarios we extracted, we designed an algorithm framework including the diffusion with transformer module, the Risk & scene encoder module, and the multimodal trajectory decoder module to predict the trajectories. In this framework, risk information and traffic flow characteristics are innovatively integrated, improving the algorithm's applicability in crash scenarios. Based on the dataset, we conducted extensive experiments. The minADE and minFDE calculated by our method are 0.009/1.159 m, which greatly improves the performance of trajectory prediction in crash scenarios. At the same time, we conducted algorithm performance tests from three aspects: the role of different modules, the complexity of the MLP linear layer in the decoder, and the influence of different modalities on the trajectory prediction results. The experimental results prove the effectiveness of the DiT module and multimodal trajectory prediction for long-tail data trajectory prediction. At the same time, it is proved that for our dataset, only a decoder with fewer linear layers is suitable. In addition, the experimental results clearly show that different long-tail data distributions will affect the prediction ability of the model. The data at the tail is often more difficult to predict because the shape of the tail data trajectory is more tortuous. Our work expands the research on long-tail trajectory prediction from both the data and the trajectory prediction framework, which can help subsequent researchers better understand the long-tail trajectories in crash scenarios.

Although the work of this paper has made good progress, the following shortcomings still exist: Firstly, since we predicted the long-tail trajectories in crash scenarios and do not use the currently popular trajectory prediction datasets such as nuScenes (*50*), Waymo (*51*), etc., it cannot be compared with the existing research results. Subsequent studies will improve the relevant models and conduct comparative experiments according to the characteristics of our dataset. Lastly, due to the limited crash data, we only used the trajectory data in the rear-end crash on the highway. Subsequent studies will continue to collect the trajectory data in crash scenarios to improve the generalization ability of the trajectory prediction model.

## CONCLUSION

Aiming at the problem of low prediction accuracy of autonomous vehicles in long-tail trajectory prediction, we extracted the trajectory data in crash scenarios containing a large number of long-tail trajectories and fused risk information such as ITTC and traffic flow characteristics as input data. At the same time, the RI-DiT trajectory prediction framework was proposed, and the diffusion with transformer module fused with risk information was adopted as the generator of trajectory features to achieve accurate prediction of long-tail trajectories. A large number of experimental results verified the importance of the DiT module and multimodal trajectories we adopted for trajectory prediction. By adding these two modules, minADE and minFDE could be reduced to 0.009/1.159 m. Meanwhile, our results show that trajectories at the tail are more difficult to predict due to their non-smooth trajectories. These results expand new ideas for long-tail trajectory prediction. Subsequent studies can help autonomous vehicles have better prediction effects in critical scenarios by fusing traffic risk information and traffic flow information.

**DECLARATION OF COMPETING INTEREST**

The authors declare that they have no known competing financial interests or personal relationships that could have appeared to influence the work reported in this paper.